\documentclass[letterpaper, 10 pt, conference]{ieeeconf}

\IEEEoverridecommandlockouts                              

\usepackage{color}

\usepackage{stackengine}

\newcommand*{\mb}[1]{{\mathbf{#1}}}

\newcommand{\dquotes}[1]{``#1''}

\newcommand{\specialcell}[2][c]{%
\begin{tabular}[#1]{@{}c@{}}#2\end{tabular}}

\usepackage{cite}
\usepackage{amsmath}
\usepackage{amssymb}
\usepackage{hyperref}
\usepackage{graphicx}

\usepackage{dsfont}
\usepackage{tikz}

\title{\LARGE \bf LaB-CL: Localized and Balanced Contrastive Learning for improving parking slot detection}

\author{U Jin Jeong, Sumin Roh$^{\dagger}$, and Il Yong Chun$^{\dagger}$
\thanks{
$^{\dagger}$Corresponding authors.
The work of U.~J.~Jeong, Sumin Roh, and I.~Y.~Chun was supported in part by NRF grants 2022R1F1A1074546 and RS-2023-00213455 funded by MSIT, KIAT grant P0022098 funded by MOTIE, and the BK21 FOUR Project.
The work of U.~J.~Jeong was additionally supported in part by Institute of Information \& communications Technology Planning \& Evaluation (IITP) grant funded by the Korea government (MSIT) (RS-2019-II190421, AI Graduate School Support Program (Sungkyunkwan University)).
The work of I.~Y.~Chun was additionally supported in part by
IITP grant 2019-0-00421 funded by MSIT,
IBS grant R015-D1,
the KEIT Technology Innovation program grant 20014967 funded by MOTIE,
SKKU-SMC and SKKU-KBSMC Future Convergence Research Program grants,
and SKKU seed grants.}
\thanks{U Jin Jeong is with the Department of Artificial Intelligence (AI), Sungkyunkwan University, Suwon 16419, South Korea. (email: 
\href{mailto:ujin4287@skku.edu}{\tt ujin4287@skku.edu})}
\thanks{Sumin Roh is with the Department of Electrical and Computer Engineering (ECE), Sungkyunkwan University, Suwon 16419, South Korea. (email: 
\href{mailto:sumsumin@skku.edu}{\tt sumsumin@skku.edu})}
\thanks{Il Yong Chun is with the Departments of  AI, ECE, Advanced Display Engineering, and Semiconductor Convergence Engineering, and the Center for Neuroscience Imaging Research, Institute for Basic Science (IBS), Sungkyunkwan University, Suwon 16419, South Korea. (email: \href{mailto:iychun@skku.edu}{\tt iychun@skku.edu})}
}

\begin{document}

\maketitle

\begin{abstract}
Parking slot detection is an essential technology in autonomous parking systems. 
In general, the classification problem of parking slot detection consists of two tasks, 
a task determining whether localized candidates are junctions of parking slots or not,
and the other that identifies a shape of detected junctions.
Both classification tasks can easily face biased learning toward the majority class, degrading classification performances.
Yet, the data imbalance issue has been \emph{overlooked} in parking slot detection.
We propose the first supervised contrastive learning framework for parking slot detection, Localized and Balanced Contrastive Learning for improving parking slot detection (LaB-CL).
The proposed LaB-CL framework uses two main approaches.
First, we propose to include class prototypes to consider representations from all classes in every mini batch, from the \emph{local} perspective.
Second, we propose a new hard negative sampling scheme that selects local representations with high prediction error.
Experiments with the benchmark dataset demonstrate that the proposed LaB-CL framework can outperform existing parking slot detection methods.
\end{abstract}

\section{Introduction}
\label{sec:intro}

Autonomous parking is a key technology in autonomous driving.
The parking slot detection is essential in autonomous parking and many recent parking slot detection methods use an around-view image \cite{DeepPS, DMPR-PS, AGNN-PSD} as they are easily accessible in modern vehicles.
The traditional parking slot detection approaches are sensitive to environmental changes \cite{wang2014automatic, jung2006parking, suhr2013full, li2017vision, suhr2013sensor}, 
so it is challenging to apply them in the real world. 
With advances in deep learning technologies,
several deep learning-based parking slot detection methods have been recently proposed \cite{DeepPS, DMPR-PS, Context_Based_PSD, AGNN-PSD}.
The deep learning-based methods \cite{DMPR-PS, AGNN-PSD} show that they are more robust to environmental changes than traditional methods \cite{wang2014automatic, jung2006parking, suhr2013full, li2017vision, suhr2013sensor}.
The predominant deep learning-based approach detects the junctions of parking slots and then uses the pairs of detected junctions to determine the location of the parking slots \cite{DeepPS, DMPR-PS, AGNN-PSD}.

The junction-based parking slot detection approach localizes and classifies junctions of parking slots.
For accurate parking slot detection,
it is crucial to accurately classify localized junction candidates.
In general, the classification problem consist of two tasks,
a task determining whether localized candidates are junctions of parking slots or not (i.e., background),
and the other that identifies the shape of detected junctions, particularly into the \dquotes{T} and \dquotes{L} shapes.
The both classification tasks can easily face the data imbalance issue, biased learning toward majority class(es) from imbalanced dataset, i.e., a dataset with  unequal distribution of classes \cite{Learning_from_imbalanced_data}, 
which can degrade classification performances.
In the first junction existence identification task,
there exist a limited number of junctions in an image, while the background class appears much more frequently. 
In the latter shape classification task, 
\dquotes{T}-shaped junctions appear much more often than \dquotes{L}-shaped junctions because L-shaped points are at the end of parking slots.
Yet, the data imbalanced issue has \emph{not} been studied in parking slot detection.

\begin{figure}[!tp]
\centering
\includegraphics[width=2.7in,clip]
{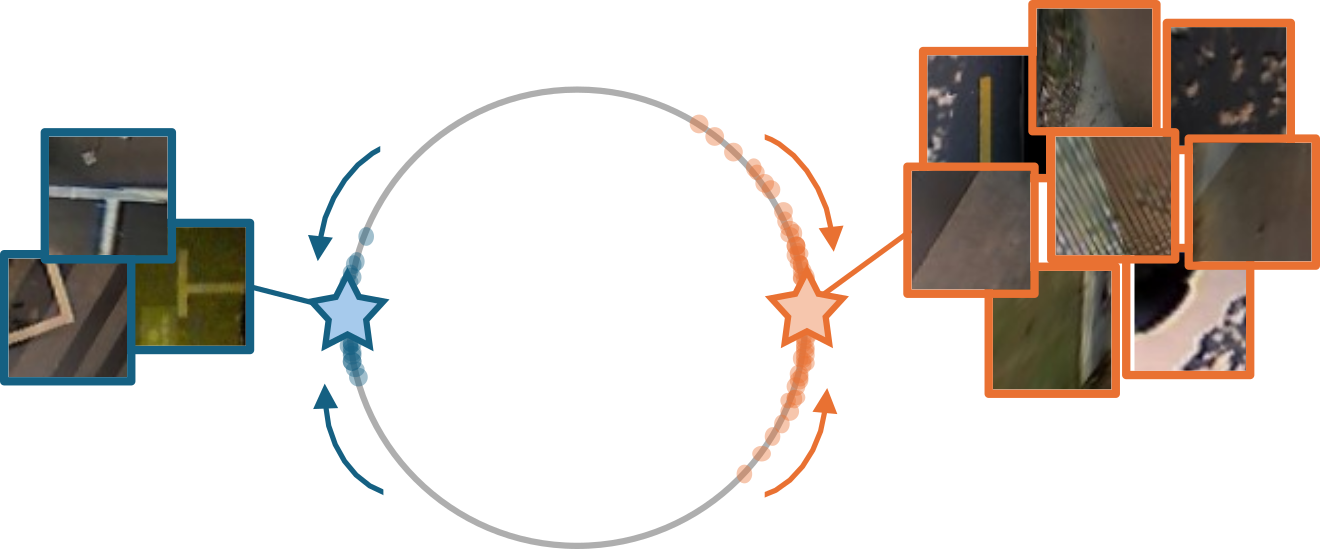}
\caption{
{\bfseries Illustration of the proposed LaB-CL framework for parking slot detection.}
The $\bullet$ denotes \emph{local} representations and the $\bigstar$ denotes a \emph{local} class prototype for each class,
and different colors mean that that instances and class prototypes are from different classes.
Proposed LaB-CL moves all the within-class local representations toward their local prototypes that are equidistant between each other while treating all classes equally.
}
\label{fig:concept}
\end{figure}

Contrastive learning (CL) is a promising representation learning approach in various applications such as image classification \cite{MoCo, SimCLR, MoCo_v2, BYOL, SimCLRv2, SCL, DetCo}, object detection \cite{MoCo, MoCo_v2, BYOL, DetCo}, and image segmentation \cite{Cross_image_CL_Seg, Contrastive_learning_semi_seg}.
The key idea of supervised CL is attracting the features of samples from same class and repulsing the features of samples from different classes.
Recently, supervised CL methods \cite{KCL, TSC, BCL} have been proposed to address the data imbalance issue in image classification.
To the best of our knowledge, supervised CL with an emphasis on imbalanced datasets has \emph{not} been investigated for object detection, particularly parking slot detection.

Many supervised CL methods learn the representations from images in full view \cite{SCL, KCL, TSC, BCL}, i.e., global representations.
In object detection problems, 
one needs to learn \emph{localized} representations because an image in full view can contain multiple objects, e.g., in the parking slot detection problem, multiple junctions of parking slots.
We propose the first supervised CL framework to improve the junction-based parking slot detection performances, {\bfseries  Localized and Balanced CL (LaB-CL)}.
Our main contributions can be summarized as follows:
\begin{itemize}
\item We propose to contrast \emph{local} representations that correspond to patterns in patches in the original image space, in supervised learning way.
For learning with imbalanced datasets, we include local prototypes in every mini batch
to consider local representations from all classes.
\item We propose a new  negative sampling scheme that selects
local representations with high prediction error.
It is possible to use the predicted probability of classes in our framework, as we simultaneously learn local representations and a classifier.
\item Experiments with the Tongji Parking-Slot (PS) benchmark dataset 2.0 (PS2.0) \cite{DeepPS} show that
the proposed Lab-CL framework outperforms existing state-of-the-art (SOTA) parking slot detection methods.
\end{itemize}

Section~\ref{sec:Related_Work} reviews related works to ours in parking slot detection and supervised CL.
Section~\ref{sec:Methods} proposes the LaB-CL framework that consists of several major components, 
new CL for junction identification and junction shape identification tasks, new hard negative sampling, and a compensation loss for CL.
Section~\ref{sec:Experimental_Results_and_Discussion} compares the proposed LaB-CL with existing parking slot detection methods and includes the ablation study on the primary components of LaB-CL.

\section{Related Work}
\label{sec:Related_Work}
\subsection{Parking slot detection}

In traditional vision-based parking slot detection methods, researchers used line and points in a hand-crafted way.
The line-based approach first detects the edge of the line with an analytical edge detector and then identifies the parking line with a line fitting algorithm such as Radon and Hough transforms \cite{wang2014automatic, jung2006parking, hamada2015surround}.
The point-based approach detects junction points of parking slots by decision tree/corner detector and then predict parking slots with template matching \cite{li2017vision, suhr2013full, suhr2013sensor}.

Recently, researchers have proposed several parking slot detection methods that use deep learning with the junction detection perspective.
Deep Parking-Slot detection (DeepPS) is the first junction-based parking slot detection method using deep learning, with two stages \cite{DeepPS}.
DeepPS first detects junctions in parking slots with a convolutional neural network (CNN) and then identifies parking slots with template matching.
Directional Marking-Point Regression (DMPR) is a two-stage method that first detects junctions using a CNN and then identifies parking slots with some post processing, specifically, manually designed geometric rules \cite{DMPR-PS}.
Attentional Graph Neural Network (AGNN) is an end-to-end (E2E) parking slot detection method with graph neural networks with attention mechanism \cite{AGNN-PSD}.

\subsection{Supervised CL for handling data imbalance}

Recently, several supervised CL methods have been proposed to address the data imbalance issue \cite{KCL, TSC, BCL}.
The $k$-positive CL method is a two-stage method that uses positive samples with a fixed number $k$ to balance between majority and minority classes \cite{KCL}.
Targeted supervised CL learning is a two-stage method that generates targets to lead all classes to have a uniform distribution in the representation space  \cite{TSC}. 
Balanced CL is an E2E method that proposes a balanced CL loss that include class prototypes from images in full view to consider all class samples in every mini batch \cite{BCL}.
Both \cite{TSC} and \cite{BCL} promote a regular simplex configuration in the representation space with imbalanced datasets.

All the aforementioned supervised CL methods learn the representations from images in full view, i.e., learn global representations.
Different from this, the proposed LaB-CL framework that can learn \emph{local} representations for junction detection of parking slots, 
while handling the data imbalance issue.
The local representation concept is similarly used in fast two-stage object detectors \cite{fast-rcnn,faster-rcnn}, but \emph{not} in supervised CL.

\begin{figure}[t!]
\centering
\small
\begin{tabular}{ccc}

\includegraphics[width=0.75in]{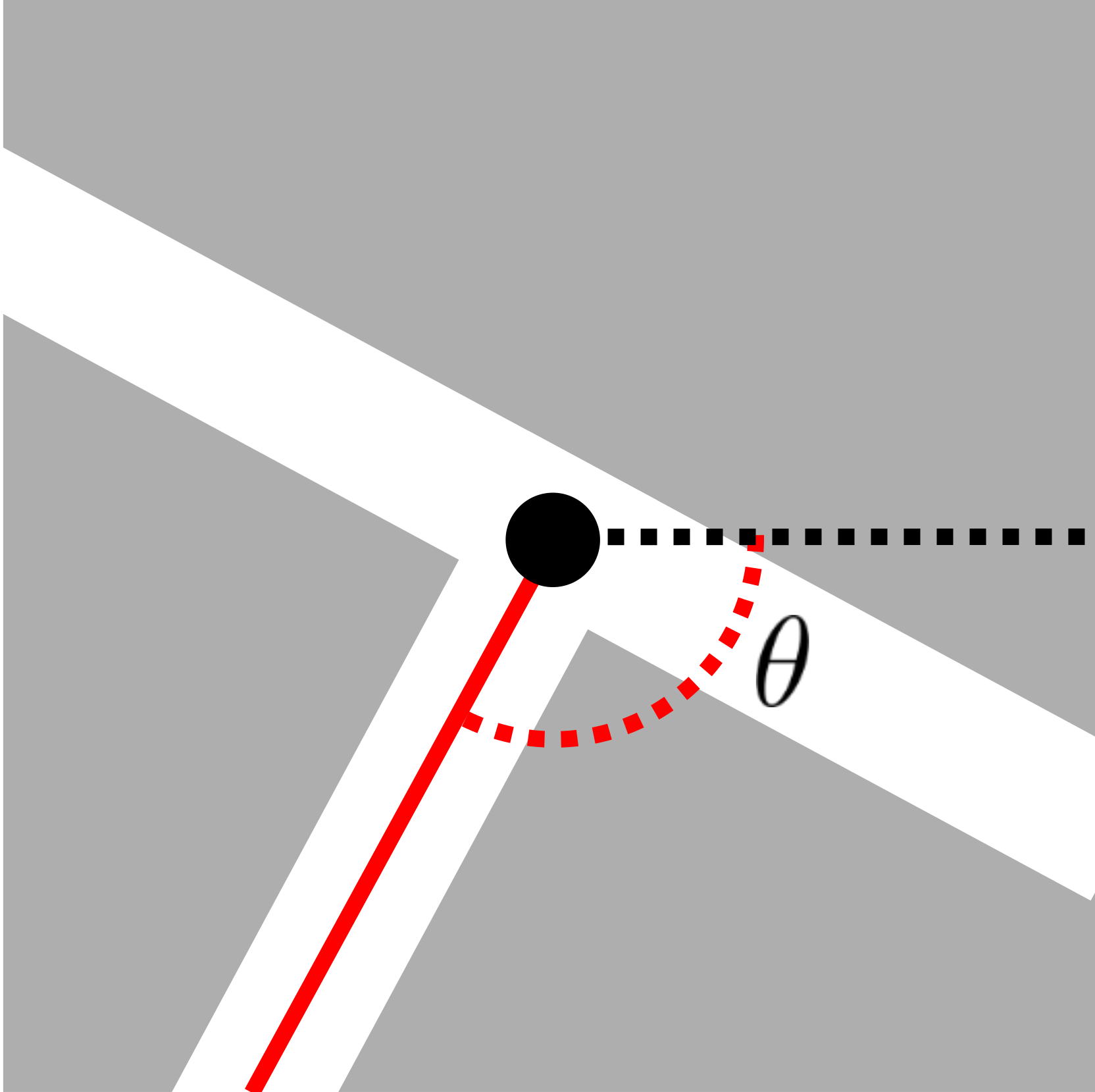}
&
\includegraphics[width=0.75in]{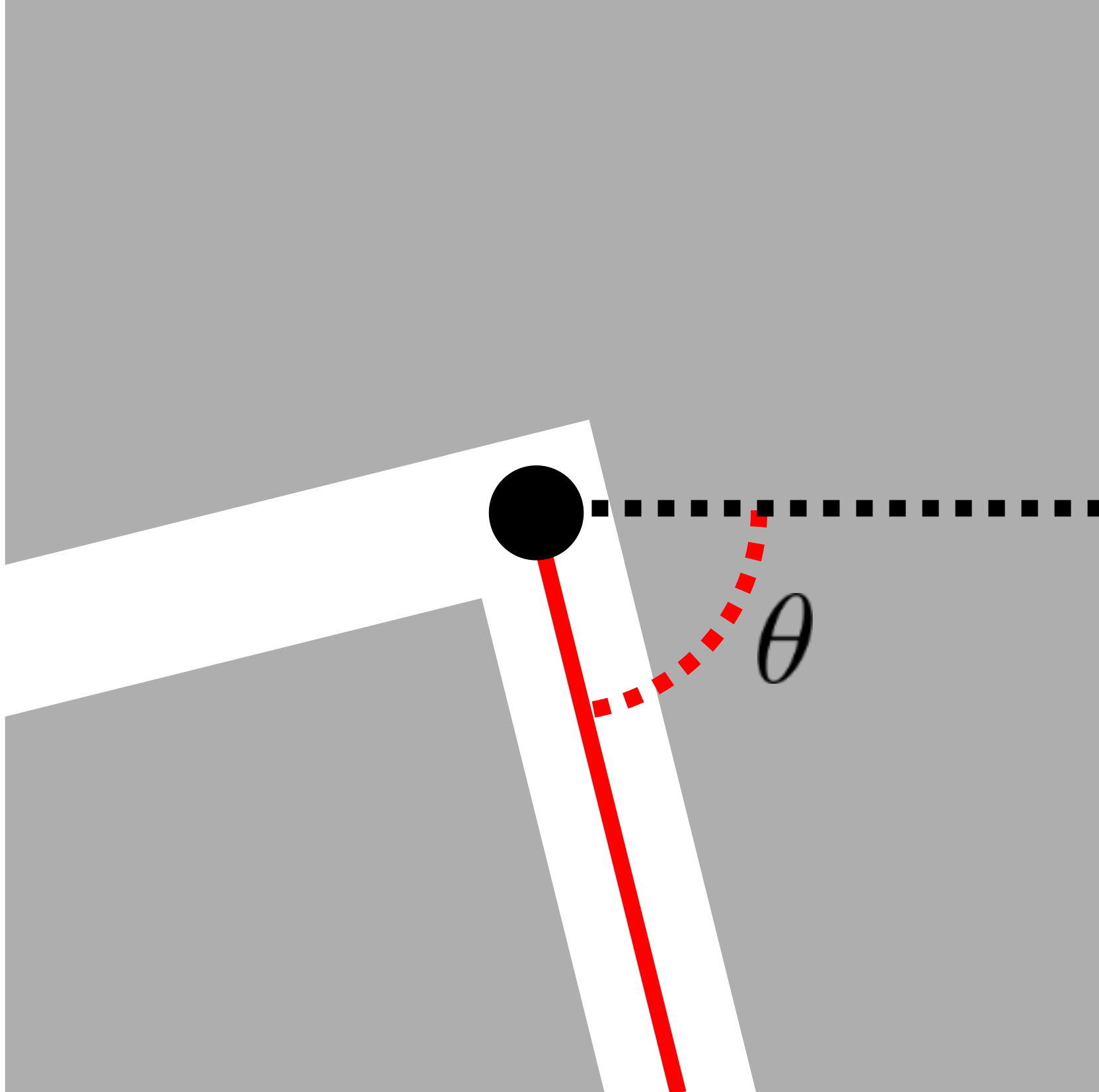}
\\
(a) T-shaped junction
& 
(b) L-shaped junction
\end{tabular}
\caption{{\bfseries Examples of junctions in parking slots.}
The black dot denotes the junction point in parking slots that is parameterized by its $x$- and $y$-coordinates.
The red dotted line denotes the rotational position of each junction that is parameterized with an angle $\theta$ from the zero angle in radians.
}
\label{fig:Marking_Point_Shape}
\end{figure}

\section{Methods}
\label{sec:Methods}

This section proposes the LaB-CL framework for accurate parking slot detection.
Section~\ref{subsec:Preliminaries} provides some preliminaries in parking slot detection.
Section~\ref{subsec:Overview} overviews the proposed LaB-CL framework and describes its main components.
Section~\ref{subsec:Proposed_CL} provides details of LaB-CL for the two classification tasks in junction detection (see Section~\ref{sec:intro}).
Section~\ref{subsec:Marking_point_detection} describes training loss for two regression tasks.
Section~\ref{subsec:Parking_slot_inference} describes postprocessing scheme that detects parking slots from detected junctions.

\begin{figure}[t!]
\centering
\small
\begin{tabular}{ccc}
\multicolumn{2}{c}{
\includegraphics[width=1.8in]{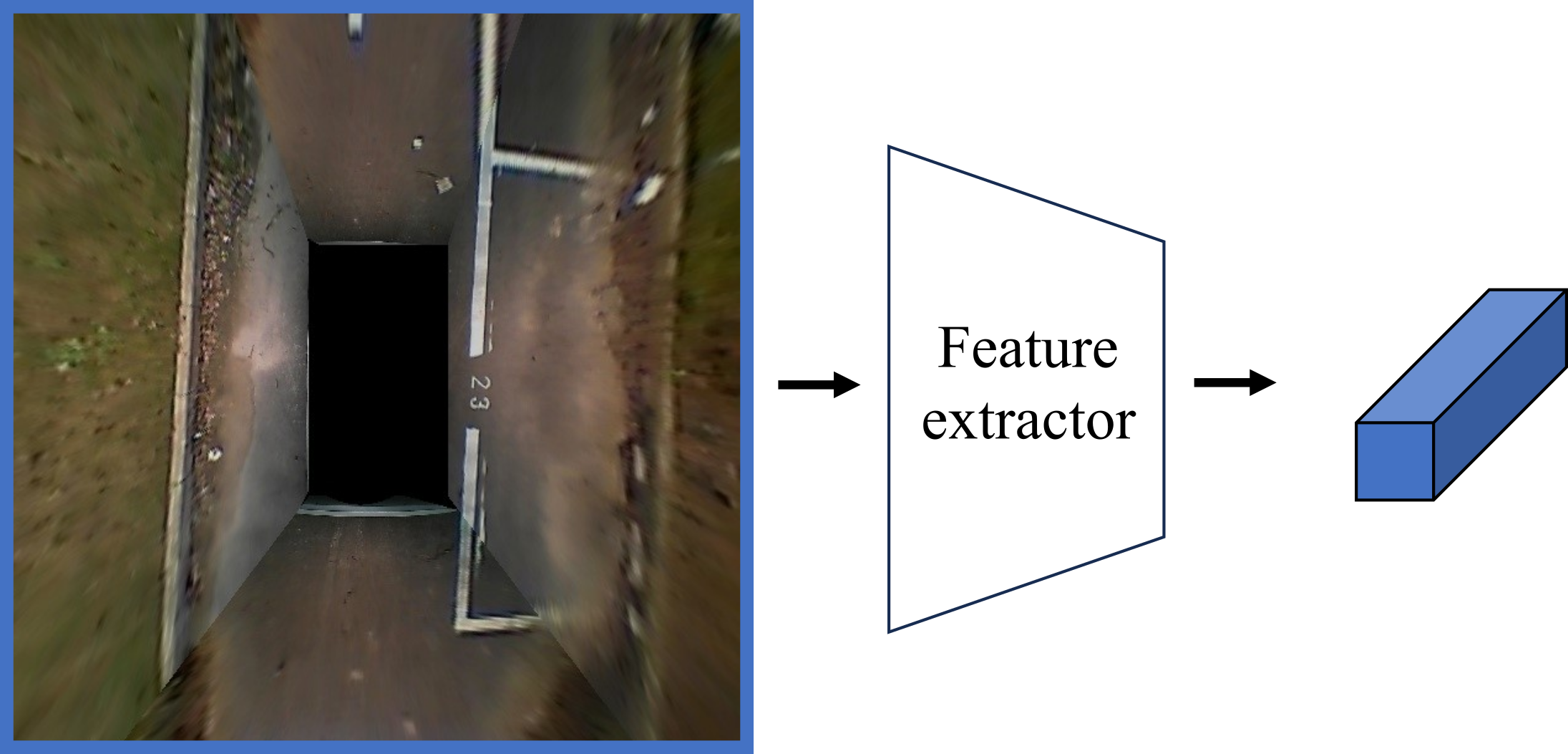}
}
\\
\multicolumn{2}{c}{(a) Global representation}
\\
\includegraphics[width=2.0in]{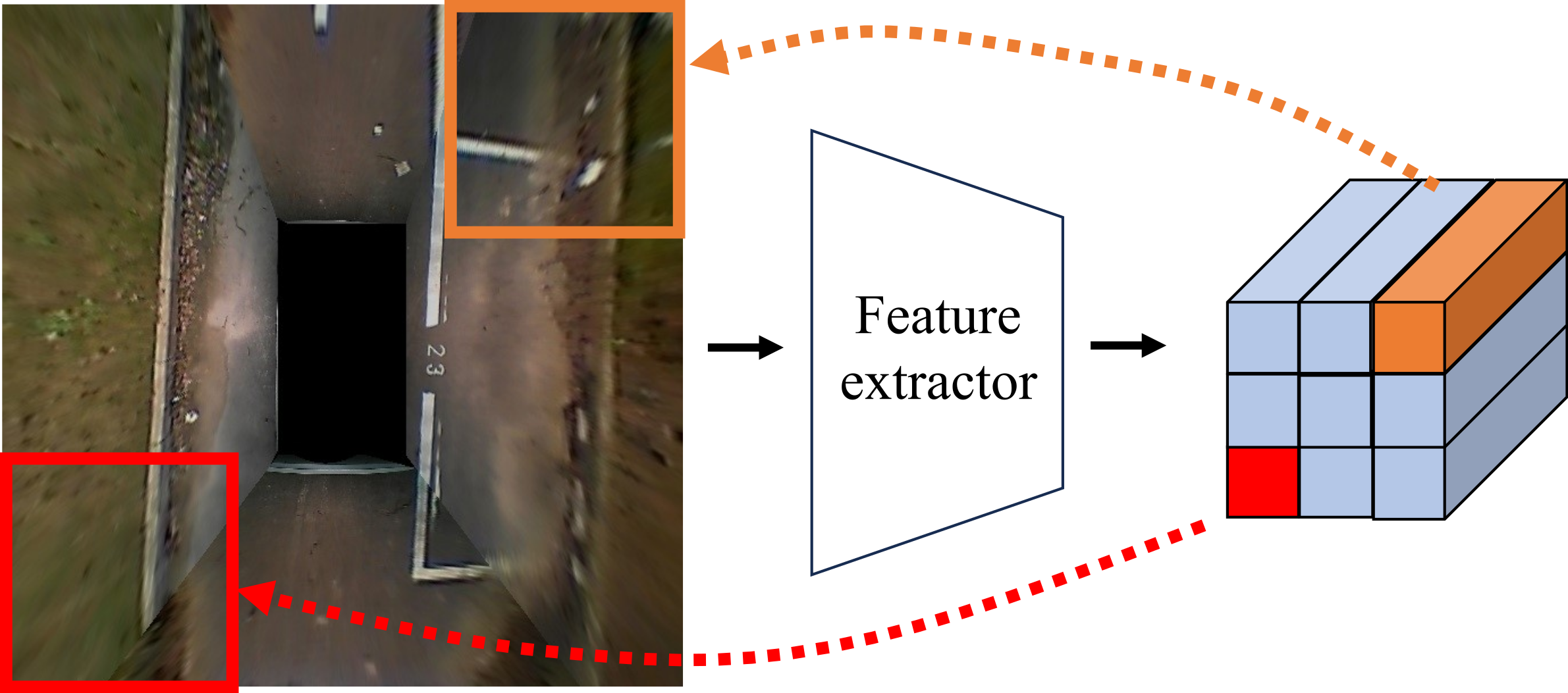}
\\
(b) Local representations
\end{tabular}
\caption{{\bfseries Illustration of the global and local representations.
(a) The global representation is extracted from the entire image. 
(b) The local representations extracted from the entire image correspond to patches in the original image space.}}
\label{fig:Global_Local_reps}
\end{figure}

\begin{figure*}[!tp]
\centering
\small
\includegraphics[width=6.2in,clip]
{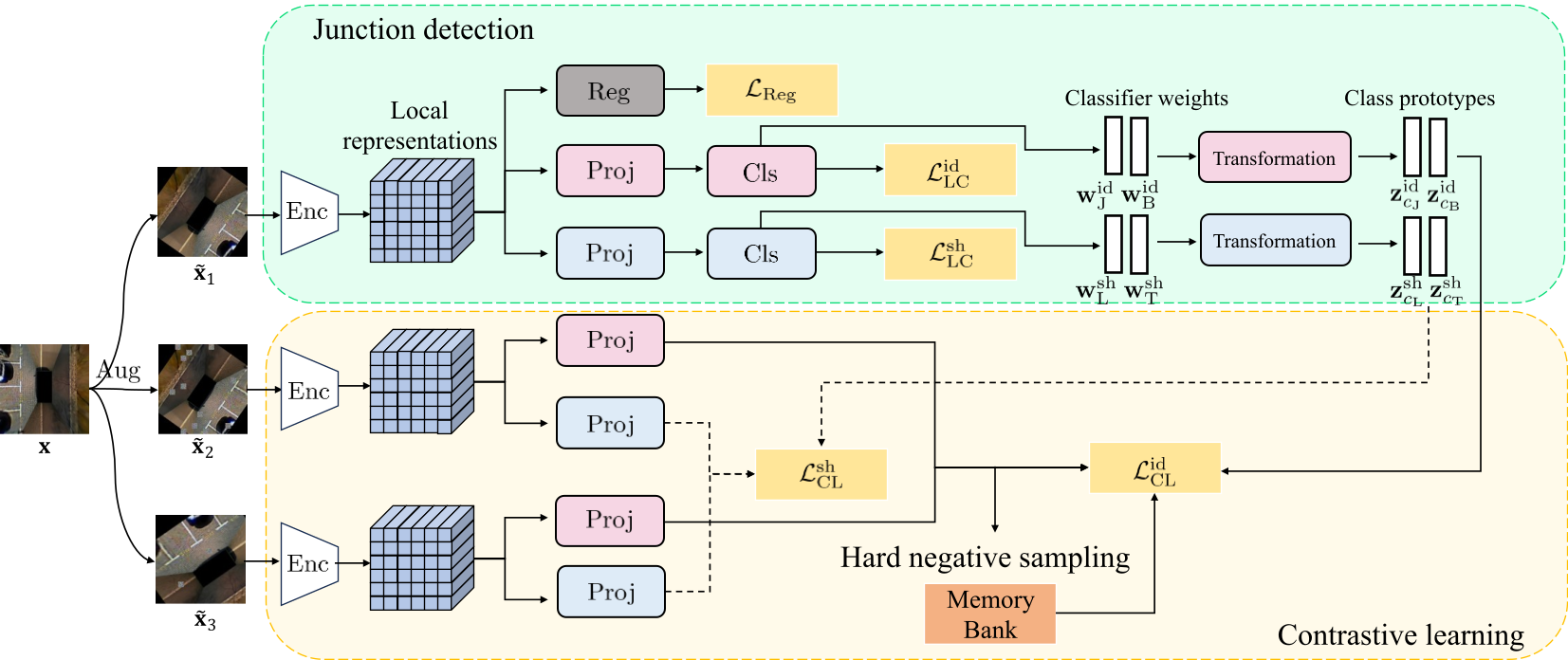}
\caption{{\bfseries Overview of the proposed LaB-CL framework for junction detection of parking slots.}
The proposed framework simultaneously learns \emph{local} representations and detector in an E2E learning manner.
{\bfseries (Localized CL)} 
We propose to use local representations rather than global representations; see Section~\ref{subsec:Preliminaries}.
{\bfseries (Balanced CL)}
We use transformed classifier weights as localized class prototypes.
In addition, we use memory banks with the proposed hard negative sampling strategy.
We include these samples in every minibatch to consider samples from all classes, i.e., balanced CL. 
{\bfseries (Postprocessing for parking slot detection)}
Some postprocessing schemes are applied to detected junctions to identify pairs of junctions and detect parking slots \cite{DMPR-PS}.}
\label{fig:Overview}
\end{figure*}

\subsection{Preliminaries}
\label{subsec:Preliminaries}

The parking slot detection task aims to identify a parking slot by detecting a pair of two junctions that constitute the entrance line of a parking slot. 
Following the earlier works \cite{DeepPS, DMPR-PS}, 
one can first detect junctions of parking slots and then detect parking slots from detected junctions with some postprocessing.

A parking slot can be parameterized by four junctions, where each junction is parameterized by its location, i.e., $x$- and $y$-coordinates, rotation angle, and a type of junction shapes \cite{DMPR-PS}.
See illustrations of the rotation angle definition depending a shape of junctions in Fig.~\ref{fig:Marking_Point_Shape}.
A junction detection model predicts the above parameters of junctions.

Fig.~\ref{fig:Global_Local_reps} illustrates the global and local representations. 
The global representation is widely used in CL \cite{MoCo, SimCLR, MoCo_v2, BYOL, SimCLRv2, SCL, DetCo, SCL, KCL, TSC, BCL}.
For CL in junction detection, we use the local representations that correspond to patches in the original image space.
Note that we input the entire image and extract local representations by modeling that each local feature vector corresponds to each patch in the original image space.
This is much faster than dividing the input image into patches and extracting features from all the patches.
The local representation concept is similarly used in fast two-stage object detectors \cite{fast-rcnn,faster-rcnn}.

\subsection{Overview}
\label{subsec:Overview}

Our goal is to develop an advanced supervised CL approach to accurately detect junctions of parking slots.
The proposed LaB-CL framework is an E2E learning method for junction detection.
Specifically, we simultaneously learn \textit{a)} feature extractors, \textit{b)} two classifiers that identify junctions and classify their shape, and \textit{c)} two regressors that predict the location and rotational position of junctions.
Mathematically, LaB-CL trains a network that maps the input space $\mathcal{X}$ to the four target spaces"
\textit{1)} the space of junction identification $\mathcal{Y^{\text{id}}}=\{\text{J}, \text{B}\}$ where J and B denote junction and background respectively;
\textit{2)} the space of junction shape classification $\mathcal{Y^{\text{sh}}}=\{\text{L}, \text{T}\}$;
\textit{3)} the space of relative junction location in a cell $\{ (x,y) \in \mathbb{R}^2 | x,y \in [0,1] \}$; and
\textit{4)} the space of rotational position of junction $\mathbb{R} \in [0, 2\pi]$.
To ultimately detect parking slots, we apply some postprocessing \cite{DMPR-PS} to detected junctions.

The proposed LaB-CL framework consists of the junction detection branch and the CL branch that learns good representations for two classification tasks. See Fig.~\ref{fig:Overview}.
LaB-CL consists of the following components, where each component is illustrated in Fig.~\ref{fig:Overview}:

\begin{itemize}

\item {\bfseries Data augmentation with three views.} 
For each input sample $\mathbf{x}$, we use random augmentation, $\tilde{\mathbf{x}} = \mathrm{Aug}(\mathbf{x})$, to generate three different views, $\{ \tilde{\mathbf{x}}_1, \tilde{\mathbf{x}}_2, \tilde{\mathbf{x}}_3 \}$. 
In the junction detection branch, we use $\tilde{\mathbf{x}}_1$.
In the CL branch, we use $\tilde{\mathbf{x}}_2$ and $\tilde{\mathbf{x}}_3$.

\item {\bfseries A feature encoder.}
An encoder extracts a set of local representations from the augmented input: $\mathbf{f} = \mathrm{Enc}(\tilde{\mathbf{x}})\in\mathbb{R}^{G \times G \times D}$, where $G$ and $D$ denote the grid size of $\mathbf{f}$ and the size of local representation vectors, respectively.
The larger $G$, the smaller patches in the original image space.
The encoder $\mathrm{Enc}$ with the same parameter is commonly used for all $\{ \tilde{\mathbf{x}}_1, \tilde{\mathbf{x}}_2, \tilde{\mathbf{x}}_3 \}$ in the junction detection and CL branches.

\item {\bfseries Two projectors and local representations in CL.}
A projector $\mathrm{Proj}$ transforms a local representation vector into the CL space.
In particular, we use different projectors for different classification tasks.
For junction identification, 
a projector produces $\mb{z}^{\text{id}}$.
For junction shape classification, 
a projector produces $\mb{z}^{\text{sh}}$.

\item {\bfseries Two classifiers.}
We denote the junction identification network and shape classification network as $\mathrm{Cls}^{\text{id}}$ and $\mathrm{Cls}^{\text{sh}}$, respectively.
They are linear classifiers with class-specific weights $\{ \mathbf{w}_{k}^{\text{id}} : k \in \mathcal{Y}^{\text{id}} \}$ and $\{ \mathbf{w}_{k}^{\text{sh}} : k \in \mathcal{Y}^{\text{sh}} \}$, respectively, and the bias.

\item {\bfseries A regressor.} 
A regression network $\mathrm{Reg}$ predicts the location and rotational position of detected junctions by giving an output vector of size $4$ for each local representation. 
Its architecture consists of $1 \times 1$ convolutional layers.
As we do not use CL for the regression tasks, we do not use projectors for them.

\item {\bfseries Local prototypes.} 
The local prototypes $\{ \mathbf{z}_{k} \}$ are transformed class-specific weights of a classifier $\{ \mathbf{w}_{k} : \forall c \}$.
The local prototypes for junction identification CL are denoted by $\{ \mathbf{z}^{\text{id}}_{c_{k}^{\text{id}}} : k \in \mathcal{Y}^{\text{id}} \}$.
The local prototypes for junction shape classification CL are denoted by $\{ \mathbf{z}^{\text{sh}}_{c_{k}^{\text{sh}}} : k \in \mathcal{Y}^{\text{sh}} \}$.

\item {\bfseries Memory banks.}
We use the memory bank $\mathcal{M}_{\text{J}}$ and $\mathcal{M}_{\text{B}}$ for the junction identification task.
\end{itemize}

\subsection{Proposed LaB-CL for parking slot detection}
\label{subsec:Proposed_CL}

We define two different local representation spaces for CL for the two different classification tasks, junction identification and shape classification.
Since we use two different local representation spaces for CL, 
we use an independent projector and classifier for each local representation space.
See Section~\ref{subsec:Overview}.

\subsubsection{Proposed CL loss for shape classification}
\label{sec:cl-sh}

We modify the CL loss from the global representation learning perspective \cite{BCL} to learn local representations as follows:
\begin{equation}
\label{eq:labcl_shape}
\begin{aligned}
    \mathcal{L}^{\text{sh}}_{\text{CL}}
    =&
    \sum_{k \in \mathcal{Y}^{\text{sh}}}
    \sum_{i \in \mathcal{J}_k} 
    \frac{-1}{| \mathcal{J}_k  |}\sum_{p\in \{ \mathcal{J}_{k}\setminus \{ i \} \} \cup \{ c^{\text{sh}}_{k} \} }  
    \\
    &
    \log\frac{\exp(\mathbf{z}^{\text{sh}}_{i}\!\cdot\! \mathbf{z}^{\text{sh}}_{p}/\tau)}{
    \sum\limits_{k'\in \mathcal{Y}^{\text{sh}}}\!\! \frac{1}{| \mathcal{J}_{k'}|+1}
    \sum\limits_{j\in \mathcal{J}_{k'}\cup \{ c^{\text{sh}}_{k'} \}}\!\!\exp(\mathbf{z}^{\text{sh}}_{i}\!\cdot\! \mathbf{z}^{\text{sh}}_{j}/\tau)},
\end{aligned}
\end{equation}
where $\mathcal{J}$ is the set of junctions of parking slots in a minibatch, 
$\mathcal{J}_k$ is a subset of $\mathcal{J}$ that includes all junction samples of $k$-shape, 
and $c^{\text{sh}}_{k}$ is the index of local prototype of junctions with $k$-shape,
for $k \in \mathcal{Y}^{\text{sh}}$.
Here, $\tau$ denotes the scalar temperature hyperparameter, and
$|\cdot|$ denotes the number of samples in a set. 
Remind that $\mathbf{z}^{\text{sh}}_i$ denotes a local representation vector for the anchor, i.e., patch, for shape classification of junctions.

The proposed loss (\ref{eq:labcl_shape}) attracts local features of the same class and repulses local features of the other class.
(\ref{eq:labcl_shape}) includes local prototypes $\{ \mathbf{z}^{\text{sh}}_{c_{k}^{\text{sh}}} : k \in \mathcal{Y}^{\text{sh}} \}$ in every mini batch during training.
Even if samples from certain classes are not included in some mini batches, including class prototypes allows a model to handle samples from both classes.
In addition, the samples of each class are averaged to ensure an equal contribution for both classes. 
Using these, one can prevent
a model is biased towards the majority class, class L.

We obtain the local prototypes by applying a non-linear multi-layer perceptron (MLP) model to the classifier weights.
This is similarly used in the other classification task, junction identification in the next section.

\subsubsection{Proposed CL loss for junction identification}
\label{sec:cl-id}

The junction identification suffers from extremely imbalanced learning, as the number of junctions of parking slots is much smaller than that of background samples.
We further modify the proposed loss (\ref{eq:labcl_shape}) to include samples from the memory banks $\mathcal{M}_{\text{J}}$ and $\mathcal{M}_{\text{B}}$:
\begin{equation}
\label{eq:labcl_exist_MB}
\begin{aligned}
    \mathcal{L}^{\text{id}}_{\text{CL}}
    \!=\! &
    \sum_{k \in \mathcal{Y}^\text{id}}
    \sum_{i\in\mathcal{A}_k}
    \frac{-1}{| \mathcal{A}_{k}|+| \mathcal{M}_{k}|} \sum_{p\in \{ \mathcal{A}_{k}\setminus \{ i \} \}\cup \{ c^{\text{id}}_{k} \}\cup \mathcal{M}_{k}} 
    \\
    &\log\frac{\exp(\mathbf{z}^{\text{id}}_{i}\!\cdot\! \mathbf{z}^{\text{id}}_{p}/\tau)}{
    \sum\limits_{k'\in \mathcal{Y}^{\text{id}}}\!\frac{1}{| \mathcal{A}_{k'}|+| \mathcal{M}_{k'}|+1}
    \sum\limits_{j\in \mathcal{A}_{k'}\cup \{ c^{\text{id}}_{k'} \}\cup \mathcal{M}_{k'} }\!\!\!\!\!\exp(\mathbf{z}^{\text{id}}_{i}\!\cdot\! \mathbf{z}^{\text{id}}_{j}/\tau)},
\end{aligned}
\end{equation}
where $\mathcal{A}$ is the set of all local representations in a minibatch, 
$\mathcal{A}_{k}$ is a subset of $\mathcal{A}$ that includes the local representations of identification class $k$,
and $c^{\text{id}}_{k}$ is the index of local prototype from identification class $k$,
for $k \in \mathcal{Y}^{\text{id}}$.

The mechanism of the proposed (\ref{eq:labcl_exist_MB}) is similar to that of (\ref{eq:labcl_shape}).
In (\ref{eq:labcl_exist_MB}), we additionally include samples from memory banks to better balance the majority and minority classes.
The memory bank $\mathcal{M}_\text{J}$ stores junction representations and the memory bank $\mathcal{M}_\text{B}$ stores hard negative background representations obtained by proposed hard negative sampling in the next section.

\subsubsection{Proposed hard negative sampling using the predicted probability of classes}
\label{sec:neg-sampl}

A large number of negative samples is important to learn good representations via CL \cite{MoCo, MoCo_v2, NPID}.
The conventional hard negative sampling approach in CL uses some distance defined in the representation space \cite{MoCHi, CL_with_Hard_Negative}.
Different from the conventional approach, 
we propose to select hard negative samples using the predicted probabilities since the proposed framework is E2E, i.e., simultaneously learns representations and predictors.
In particular, we define a hard negative sample as a sample with a high prediction error.
In practice, 
we select the top $2\%$ hard negative samples from every mini batch and store them in $\mathcal{M}_B$.

We apply the proposed hard negative sampling scheme to construct the background memory bank $\mathcal{M}_B$,
i.e., 
we select background samples that are misidentified as junction with a high prediction error.
In the junction identification task,
junctions generally have some specific patterns, while background samples do not.
So, it is our general observation that background features can be easily misclassified as junctions, i.e.,
the number of hard negative samples in the background class is larger than that in the junction class.

\subsubsection{Further promoting attractions in CL}
\label{sec:cl-attr}

To further promote in junction identification that all within-class local representations collapse to their class means,
we use the following loss \cite{HyCon}:
\begin{equation}
\label{eq:labcl_Refinement}
\begin{aligned}
\mathcal{L}_{\text{A}}^{\text{id}} 
=& 
\sum_{k \in \mathcal{Y}^\text{id}}
\sum_{i \in \mathcal{A}_k}
\frac{1}{| \mathcal{A}_{k} | + | \mathcal{M}_k |}
\\
&
\sum_{p\in \{ \mathcal{A}_{k} \setminus \{ i \} \}\cup \{ c^{\text{id}}_{k} \} \cup \mathcal{M}_{k} }
\left\| \mathbf{z}^{\text{id}}_{i}\!\cdot\!\mathbf{z}^{\text{id}}_{p} - 1 \right\|_2^2,
\end{aligned}
\end{equation}
where all the notations are defined as in (\ref{eq:labcl_exist_MB}).
The motivation of using the loss function (\ref{eq:labcl_Refinement}) is different from  \cite{HyCon} that helps to have more stable training, 
as such unstable training issue can be resolved by (\ref{eq:labcl_exist_MB}).
We rather consider that the variability of background local representations is large by their nature.
To further promote that all local representations from the background class collapse to its means,
we propose to use (\ref{eq:labcl_Refinement}).

\subsubsection{Logit compensation losses for learning classifiers} 

The logit compensation (LC) method \cite{Logit_adjustment, BCL} aim to moderate the bias caused by data imbalance. 
The logit compensation loss function for junction shape classification is given by
\begin{equation}
\label{eq:labcl_LC_shape}
    \mathcal{L}^{\text{sh}}_{\text{LC}}
    =
    \frac{-1}{|\mathcal{J}|}\sum_{i \in \mathcal{J}} \sum_{k \in \mathcal{Y}^\text{sh}}^{} \log\frac{\exp (\mathrm{Cls}^\text{sh}(\mathbf{z}^\text{sh}_i)_{k}+\log(q^\text{sh}_{k}))}{\sum\limits_{k' \in \mathcal{Y}^\text{sh}} \exp(\mathrm{Cls}^\text{sh}(\mathbf{z}^\text{sh}_i)_{k'}+\log(q^\text{sh}_{k'}))},
\end{equation}
where $q^{\text{sh}}_{k} = |\mathcal{J}_{k}| / |\mathcal{J}|$ represents the frequency of $k$-shape, $k \in \mathcal{Y}^\text{sh}$.
The logit compensation loss function for junction identification $\mathcal{L}^{\text{id}}_{\text{LC}}$ similarly given by some modifications of sets and an operator, etc.

\begin{figure*}[t]
    \centering
    \begin{tabular}{cccccc}
    
    \begin{tikzpicture}
        \node[anchor=center, inner sep=0] at (-0.40\textwidth,0.15) {\includegraphics[width=0.15\textwidth]{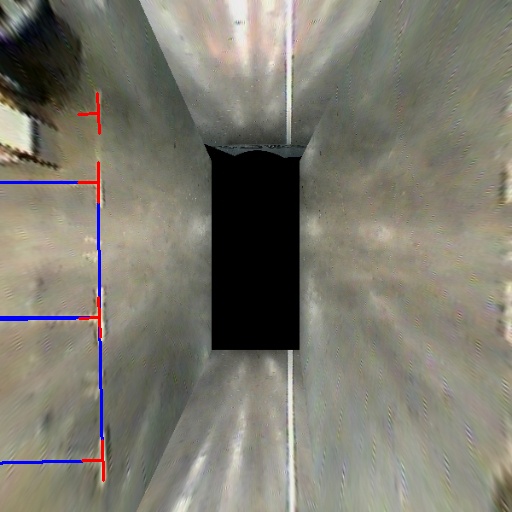}};
        
        \node[anchor=center, inner sep=0] at (-0.24\textwidth,0.15) 
        {\includegraphics[width=0.15\textwidth]{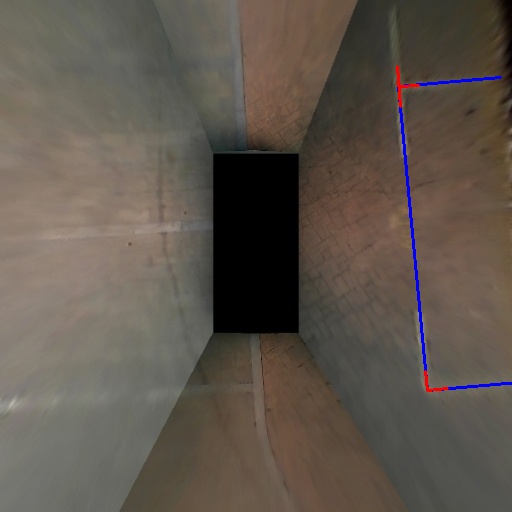}};
        
        \node[anchor=center, inner sep=0] at (-0.08\textwidth,0.15) 
        {\includegraphics[width=0.15\textwidth]{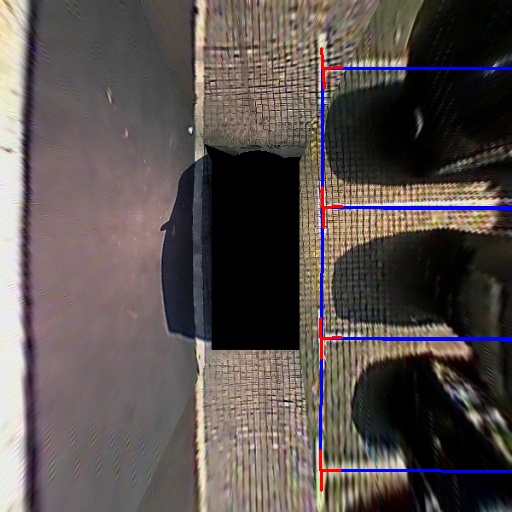}};
        
        \node[anchor=center, inner sep=0] at (0.08\textwidth,0.15) {\includegraphics[width=0.15\textwidth]{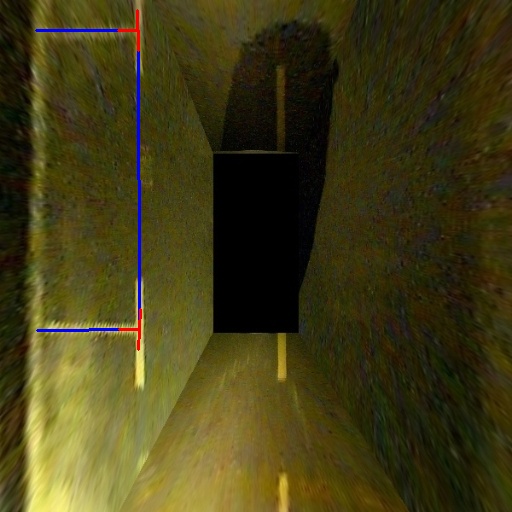}};
        
        \node[anchor=center, inner sep=0] at (0.24\textwidth,0.15) {\includegraphics[width=0.15\textwidth]{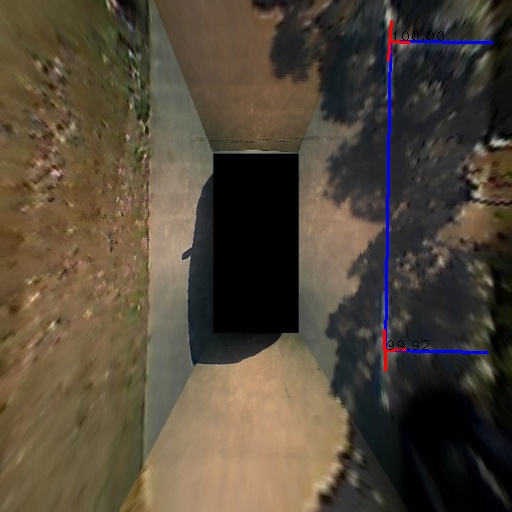}};

        \node[anchor=center, inner sep=0] at (0.40\textwidth,0.15) {\includegraphics[width=0.15\textwidth]{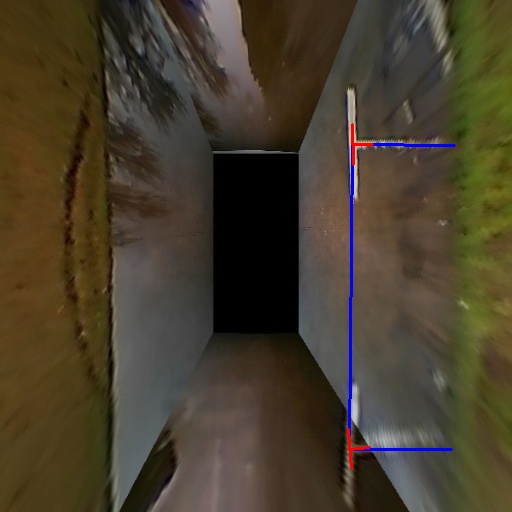}};

        \node[align=center] at (-0.40\textwidth,1.8) {\small (a) Indoor};
        \node[align=center] at (-0.24\textwidth,1.8) {\small (b) Outdoor};
        \node[align=center] at (-0.08\textwidth,1.8) {\small (b) Daytime};
        \node[align=center] at (0.08\textwidth,1.8) {\small (d) Nighttime};
        \node[align=center] at (0.24\textwidth,1.8) {\small (e) Sunny};
        \node[align=center] at (0.40\textwidth,1.8) {\small (f) Rainy};

    \end{tikzpicture}
    \end{tabular}
    \caption{{\bfseries Qualitative parking slot detection results under various conditions.}
    Red indicates the junction, and blue indicates the entrance and side lines.}
    \label{fig:parking_slot_results}    
\end{figure*}
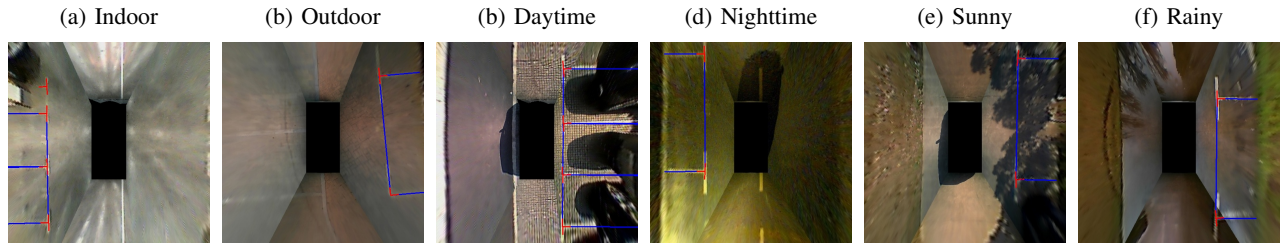

\subsection{Regression losses for learning predictors of junction location and rotation}
\label{subsec:Marking_point_detection}

For learning a regression network that predicts the location and rotation angle of junction,
we use the following sum of squared errors loss:
\begin{equation} 
\label{eq:labcl_Reg_Loss}
\begin{aligned}
    \mathcal{L}_{\text{Reg}} 
    =\,& 
    \frac{1}{|\mathcal{J}|}
    \sum_{i \in \mathcal{J}} 
    (x_i - x^\text{GT}_i)^2 + 
    (y_i - y^\text{GT}_i)^2 +
    \\
    &
    (\cos(\theta_i) - \cos(\theta^\text{GT}_i))^2 +
    (\sin(\theta_i) - \sin(\theta^\text{GT}_i))^2,
\end{aligned}
\end{equation}
where $(x_i, y_i)$ and $\theta_i$ are the predicted location and rotation angle of the $i$th junction, respectively, and 
the superscript $\text{GT}$ denotes the gound truth labels.

Finally, our overall loss function is given as follows:
\begin{equation} 
\label{eq:Overall_loss}
\begin{aligned}
    \mathcal{L} =\, &
    \lambda^{\text{sh}}_{\text{CL}} \mathcal{L}^{\text{sh}}_{\text{CL}} + \lambda^{\text{id}}_{\text{CL}}
    \mathcal{L}^{\text{id}}_{\text{CL}} + \mathcal{L}^{\text{id}}_{\text{A}} \mathcal{L}^{\text{id}}_{\text{A}} + \\
    &\lambda^{\text{sh}}_{\text{LC}} \mathcal{L}^{\text{sh}}_{\text{LC}} + 
    \lambda^{\text{id}}_{\text{LC}}
    \mathcal{L}^{\text{id}}_{\text{LC}} + \lambda_{\text{Reg}}\mathcal{L}_{\text{Reg}},
\end{aligned}
\end{equation}
where $\mathcal{L}^{\text{sh}}_{\text{CL}}$, 
$\mathcal{L}^{\text{id}}_{\text{CL}}$,
$\mathcal{L}^{\text{id}}_{\text{A}}$,
$\mathcal{L}^{\text{sh}}_{\text{LC}}$,
$\mathcal{L}^{\text{id}}_{\text{LC}}$,
and $\mathcal{L}_{\text{Reg}}$
are given as in
(\ref{eq:labcl_shape}), 
(\ref{eq:labcl_exist_MB}),
(\ref{eq:labcl_Refinement}),
(\ref{eq:labcl_LC_shape}),
the variant of (\ref{eq:labcl_LC_shape}),
and (\ref{eq:labcl_Reg_Loss}),
respectively,
and 
$\{ \lambda^{\text{sh}}_{\text{CL}}, 
\lambda^{\text{id}}_{\text{CL}},
\lambda^{\text{id}}_{\text{C}},
\lambda^{\text{sh}}_{\text{LC}},
\lambda^{\text{id}}_{\text{LC}},
\lambda_{\text{Reg}} \}$
are their hyperparameters that control the strength of each loss.

\subsection{Postprocessing for parking slot detection}
\label{subsec:Parking_slot_inference}

To eliminate duplicate detections and select the most relevant detected jections,
we apply the non-maximum suppression (NMS) technique to detected junctions -- as conventionally used in object detection. 
To ultimately detect parking slots, the following two postprocessing techniques are applied after NMS \cite{DMPR-PS}:
\textit{1)} filtering inappropriate junction pairs and \textit{2)} pairing junctions to form parking slot entrance lines.
See further details in \cite{DMPR-PS}.

\section{Experimental Results and Discussion}
\label{sec:Experimental_Results_and_Discussion}

\subsection{Experimental setups}

We compared the proposed LaB-CL framework with  traditional parking slot detection methods, 
Wang et al.~\cite{wang2014automatic}, Hamda et al.~\cite{hamada2015surround}, and PSD\_L \cite{li2017vision},
and SOTA deep learning-based parking slot detection methods with the junction detection perspective,
DeepPS \cite{DeepPS}, DMPR \cite{DMPR-PS}, and AGNN \cite{AGNN-PSD}.

\begin{table}
\caption{Comparisons of parking slot detection performances with different methods (PS2.0 dataset)}
\label{tab:table1}
\centering
\begin{tabular}{c|cc}
\hline\hline
Method    & Precision    & Recall   \\ 
\hline
Wang et al.~\cite{wang2014automatic} &98.29\% & 58.33\%  \\
Hamda et al.~\cite{hamada2015surround}& 98.45\% & 61.37\% \\
PSD\_L \cite{li2017vision}& 98.41\% & 86.96\% \\ 
DeepPS \cite{DeepPS}& 98.99\% & 99.13\% \\ 
DMPR \cite{DMPR-PS}& 99.42\% & 99.37\% \\ 
AGNN \cite{AGNN-PSD}& 99.56\% & 99.42\% \\ 
\hline
{\bfseries LaB-CL (ours)} & {\bfseries 99.57\%} & {\bfseries 99.81\%} \\ 
\hline\hline
\end{tabular}
\end{table}

\subsubsection{Dataset} 

We used the PS2.0 benchmark dataset \cite{DeepPS} that 
consists of $9,\!827$ training images and $2,\!338$ test images. 
The PS2.0 dataset includes both indoor and outdoor images collected under various environmental conditions.
Images are with the size of $600 \!\times\! 600$, corresponding to a $10$m$\times$$10$m physical region.
Following \cite{DMPR-PS, AGNN-PSD}, we used the selected $7,\!844$ images for training and $2,\!290$ images for test.

The proposed method resizes the original image resolution to $512 \!\times\! 512$ for the input,
and we set the grid size $G$ as $16$.
Consequently,
from the local representations perspective, 
the imbalance factor $\rho$ is approximately $128$ and $137$ for training and test dataset of junction identification, respectively, and
$\rho$ is approximately $7$ and $10$ for training and test dataset of junction shape classification, respectively. 
Here, $\rho = N_{\text{max}} / N_{\text{min}}$, and $N_{\text{max}}$ and $N_{\text{min}}$ is the number of samples in the majority and minority classes, respectively.

\subsubsection{Implementation details} 

Inspired by \cite{DMPR-PS, SimCLR}, we used random rotation, random resize crop, Gaussian blur, and random erasing for data augmentation.
We set $\lambda^{\text{sh}}_{\text{CL}}$,
$\lambda^{\text{id}}_{\text{CL}}$,
$\lambda^{\text{id}}_{\text{A}}$,
$\lambda^{\text{sh}}_{\text{LC}}$,
$\lambda^{\text{id}}_{\text{LC}}$, and $\lambda_{\text{Reg}}$ 
as $1$, $1$, $10$, $1$, $100$, and $1$, respectively.
The temperature $\tau$ was set to $0.1$ \cite{BCL}.
We trained the model with the batch size of $24$ with  $1,\!000$ epochs. 
We used the Adam optimizer with the initial learning rate of $0.001$, and decayed it at the $800$th and $900$th epoch with the decay rate of $0.1$.
For all the LaB-CL experiments, we used ResNet-50 \cite{ResNet} as the feature encoder $\mathrm{Enc}$.
For the two projectors, $\mathrm{Proj}$, we used the three-layer MLP architecture \cite{SimCLRv2}.
The local prototypes were transformed versions of class-specific weights with two-layer MLPs \cite{BCL}.
We set the dimensions of the hidden layer and output layer three- and two-layer MLPs as $2048$ and $512$,
respectively.
We set the size of the memory bank to $256$.

\subsubsection{Evaluation metric} 

We determine that the parking slot detection is correct if the root mean squared error between predicted locations and their ground truths is less than $10$ \cite{DMPR-PS,AGNN-PSD}.
(Note that locations of detected junctions are transformed in parking slot detection.)

\subsection{Comparisons b/w different parking slot detection methods}

\begin{figure}[t!]
\centering
\small
\begin{tabular}{ccc}

\includegraphics[width=1.25in]{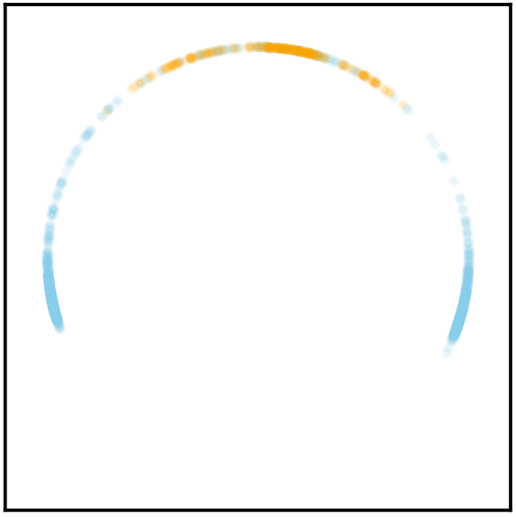}
&
\includegraphics[width=1.25in]{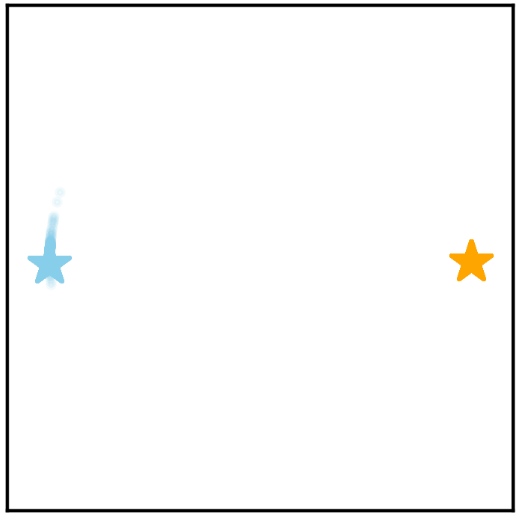}
\\
(a) DMPR
& 
(b) {\bfseries LaB-CL (ours)}
\end{tabular}
\caption{{\bfseries Visualization of local representation distribution on the unit sphere} (PS2.0 dataset; $\{ x \in [-1, 1], y \in [-1, 1] \}$). 
If within-class representations are collapsed  to form a regular simplex configuration, then good representations are learned.
The $\bullet$ denotes a local representation and $\bigstar$ denotes a local class prototype. 
The orange denotes local representations of L-shaped junctions, the skyblue denotes local representations of T-shaped junctions.}
\label{fig:Local representation distribution}
\end{figure}

\subsubsection{Parking slot detection performance comparisons}

Results in Table~\ref{tab:table1} with the PS2.0 benchmark dataset demonstrates that the proposed LaB-CL framework can outperform existing SOTA parking slot detection methods. 
In particular, proposed LaB-CL significantly improves the recall compared to existing SOTA methods.
Fig.~\ref{fig:parking_slot_results} shows the parking slot detection results obtained by LaB-CL. 
It shows that the proposed method can achieve accurate parking point detection regardless of environmental conditions.

The averaged processing/analyzing time for a single image with the proposed method is about $9.16$ms,
where we measured the inference speed with a single Nvidia GeForce RTX 4090 GPU.

\subsubsection{Representation learning performance comparisons}

The proposed LaB-CL framework can promote that within-class local features collapse to their prototypes, forming a regular simplex geometry.
Such phenomenon is not found in an existing SOTA method, DMPR.
See Fig.~\ref{fig:Local representation distribution}.
Inducing a simple and symmetric geometry in the representation space is important for improving the generalization and robustness \cite{pnas}.
In particular, with such learned representations, one can learn a simple yet effective classifier (e.g., nearest centroid classifier).

\subsection{Ablation study for proposed LaB-CL} 

Table~\ref{tab:table2} reports results from the ablation study for the primary components of proposed LaB-CL framework.
Compared with the baseline that does not use any of our innovations (in the first row of Table~\ref{tab:table2}),
the proposed LaB-CL innovation \dquotes{CL+Hard negative sampling} (in the third row of Table~\ref{tab:table2}) can significantly improve the parking slot detection performances.
Further attracting within-class local representations can further improve the parking slot detection performances of the LaB-CL.
Compare the \dquotes{CL+Hard negative sampling} combination with the \dquotes{CL+Hard negative sampling+Further attraction} combination (in the last row of Table~\ref{tab:table2}).

\begin{table}
\caption{Ablation study for the primary components of LaB-CL (PS2.0 dataset)}
\label{tab:table2}
\centering
\begin{tabular}{ccc|cc}
\hline\hline
\specialcell[c]{CL \\ 
(\S\ref{sec:cl-sh}--\\ \ref{sec:cl-id})} & 
\specialcell[c]{Hard negative \\ 
sampling \\
(\S\ref{sec:neg-sampl})} & 
\specialcell[c]{Further \\ 
attraction \\
(\S\ref{sec:cl-attr})}
 & Precision    & Recall   \\ 
\hline
$\times$ & $\times$ & $\times$ & 99.42\% & 99.37\% \\ 
$\checkmark$ & $\times$ & $\times$ & 99.47\% & 99.66\% \\ 
$\checkmark$ & $\checkmark$ & $\times$ & 99.52\% & 99.71\% \\ 
$\checkmark$ & $\times$ & $\checkmark$ & 99.52\% & 99.66\% \\ 
$\checkmark$ & $\checkmark$ & $\checkmark$ & 99.57\% & 99.81\% \\ 
\hline\hline
\end{tabular}
\end{table}

\section{Conclusion}
\label{sec:Conclusion}

Autonomous parking can reduce the need for large parking lots and spaces and alleviate traffic congestion.
To develop safe autonomous parking systems, 
it is crucial develop accurate parking slot detection models.
The predominant deep learning-based approach first detects the junctions of parking slots and then, detects parking slots with detected junctions with hand-crafted postprocessing.

The proposed LaB-CL framework is the first E2E supervised CL for junction detection. 
To moderate the data imbalance issue in junction detection, 
LaB-CL \textit{1)} includes \emph{local} class prototypes in every mini batch to consider local representations from all classes,
and \textit{2)} selects hard negative representations with high prediction error.
Lab-CL achieves outperforming parking slot detection performances compared to several existing SOTA parking slot detection methods.

We expect to further improve performances of LaB-CL by replacing hand-crafted postprocessing with a learnable module, like \cite{AGNN-PSD}.





\bibliographystyle{IEEEtran}
\bibliography{refs}

\end{document}